\documentclass[10pt]{article}
\usepackage[preprint]{tmlr}

\usepackage{hyperref}
\usepackage{url}
\usepackage{booktabs}
\usepackage{amsfonts}
\usepackage{amssymb}
\usepackage{nicefrac}
\usepackage{microtype}
\usepackage{xcolor}
\usepackage{graphicx}
\usepackage{enumitem}

\newif\ifvenueaamas
\venueaamasfalse
\newcommand{\smallfigwidth}{0.55\textwidth}

\PassOptionsToPackage{final}{changes}
\usepackage{array}
\usepackage{longtable}
\usepackage{amsmath}
\usepackage{placeins}
\usepackage{tikz}
\usepackage{fancyvrb}
\usepackage[commandnameprefix=ifneeded]{changes}
\setaddedmarkup{\textcolor{red}{#1}}
\setdeletedmarkup{\,\textcolor{black}{\sout{#1}}\,}
\setauthormarkup{}
\usetikzlibrary{arrows.meta,positioning}

\usepackage{flafter}

\title{CONFIRM: Governing Scientific Reporting for Trustworthy Agentic Neuroimaging}

\author{%
  \name Yuxiang Wei\textsuperscript{1,2},
  Lijing Zhu\textsuperscript{3},
  Qizhen Lan\textsuperscript{4},
  Yanteng Zhang\textsuperscript{1,5}, and
  Vince D. Calhoun\textsuperscript{1,2,5,6} \\[4pt]
  \addr \textsuperscript{1}TReNDS Center, Atlanta, GA, USA \\
  \textsuperscript{2}Georgia Institute of Technology, Atlanta, GA, USA \\
  \textsuperscript{3}University of Houston--Clear Lake, Houston, TX, USA \\
  \textsuperscript{4}McWilliams School of Biomedical Informatics, \\
  The University of Texas Health Science Center at Houston, Houston, TX, USA \\
  \textsuperscript{5}Georgia State University, Atlanta, GA, USA \\
  \textsuperscript{6}Emory University, Atlanta, GA, USA \\[7pt]
  {\normalfont\small\textbf{Project page:}~\url{https://yuxiangwei0808.github.io/CONFIRM/}}
}
\date{}

\hypersetup{
  pdftitle={CONFIRM: Governing Scientific Reporting for Trustworthy Agentic Neuroimaging},
  pdfauthor={Yuxiang Wei, Lijing Zhu, Qizhen Lan, Yanteng Zhang, Vince D. Calhoun},
  pdfsubject={Scientific reporting governance for agentic neuroimaging},
  pdfkeywords={LLM agents, scientific discovery, claim governance, statistical validity, reproducibility, neuroimaging, benchmark}
}

\begin{document}

\maketitle

\begin{abstract}
Large language model (LLM) agents can autonomously run neuroimaging analyses from research
questions to reported findings. Existing agentic
systems use tools to check that code runs and reported values match its output, but rely on
LLM judgment to decide which results warrant scientific reporting. We show that LLM judges endorse site-confounded findings even when the
confound is presented as evidence. We present \textbf{CONFIRM}, a framework that separates running an analysis from deciding which findings to report. CONFIRM encodes each hypothesis as a machine-readable, pre-specified analysis plan that fixes the claim and its pass criteria before computing results. It then filters results through five deterministic gates (search history and multiplicity, confounds, power, analytic stability, and replication) under three configurable reporting standards, from least to most demanding: \textit{Screen}, \textit{Replicate}, and \textit{Full}. Failed claims receive diagnoses that guide follow-up analyses. To evaluate reporting decisions at the claim level, we introduce \textbf{NeuroClaimBench}, comprising 489 executable neuroimaging
claims with frozen data, including literature-supported findings and 150 controls
designed to violate individual methodological requirements.
On NeuroClaimBench, CONFIRM falsely supports at most 3\% of controls, compared with 23--38\% for the baselines, while recovering up to 61\% of literature-supported claims. When auditing an existing agent, CONFIRM reduces false support on controls from 23\% to 0\% while retaining the agent's literature-supported claims. Its diagnostic feedback also produces more supported revisions than Self-Refine with fewer LLM calls.
Together, CONFIRM and NeuroClaimBench show that basing reporting decisions on pre-specified analysis plans and deterministic gates is essential for trustworthy autonomous scientific discovery.
\end{abstract}

\section{Introduction}
\label{sec:intro}

Large language model (LLM) agents are increasingly deployed to automate scientific discovery: given a research hypothesis, an agent selects datasets, formulates an analysis, executes code, and summarizes the findings~\citep{boiko2023}. In neuroimaging, agents have recently begun to automate this entire process~\citep{neuroclaw2026,neuroagent2026,han2026,neura2026}. For example, when testing whether hippocampal volume decreases in Alzheimer's disease, an agent can autonomously compare patients with dementia against cognitively normal controls in ADNI~\citep{weiner2010adni} and report the resulting group difference. To date, verification frameworks for these workflows focus almost entirely on checking that scripts run without error and that reported values match their outputs.

However, computational correctness does not guarantee scientific validity. A correctly computed result may still come from an underpowered cohort, an unadjusted confound, or a comparison selected after inspecting the results. These problems are common in neuroimaging: reliable brain-wide associations often require thousands of participants~\citep{marek2022}, independent analysis pipelines frequently reach different conclusions from the same raw data~\citep{botviniknezer2020}, and small cohorts routinely inflate effect sizes that fail to replicate~\citep{button2013,poldrack2017}.

Current systems entrust the decision to report a finding to the LLM itself, which acts as an \emph{LLM judge}. LLMs can report values unsupported by their tools \citep{ji2023}, misjudge the applicability of statistical tests~\citep{zhu2024statqa}, and endorse site-confounded results even when shown evidence of the confound (Section~\ref{sec:ncb_v21_results}). In these cases, the compared groups come mostly from different scan sites, so site differences can appear as group effects. Fixed heuristics, such as filtering for $p<0.05$, offer little protection because significance thresholds accept effects caused by confounds just as readily as genuine effects. Spurious findings distort downstream meta-analyses and power calculations, and direct follow-up studies toward effects that fail to replicate.

We introduce \textbf{CONFIRM}, a framework that separates running an analysis from deciding whether its finding warrants scientific reporting. We call this decision process \emph{claim governance}. The LLM first translates a hypothesis into a \emph{pre-specified analysis plan} that fixes the estimand (the effect being tested), covariates (variables included to adjust for other influences), cohorts, search family (every test the agent may run for the question), and pass criteria. Deterministic gates are fixed rules that evaluate five properties: search history and multiplicity (whether the result survives correction for every test run), measured confounds, statistical power (the probability of detecting a specified effect), analytic stability (whether at least 60\% of reasonable alternative analyses agree), and independent replication. The \textit{Screen}, \textit{Replicate}, and \textit{Full} reporting standards specify which gates a claim must pass. CONFIRM runs as a standalone agent or audits an existing agent in \emph{retrofit mode}. Failed-gate diagnoses guide follow-up analyses under the pre-specified analysis plan.

Existing benchmarks assess program execution or agreement with reference code~\citep{chen2025scienceagentbench}. We introduce \textbf{NeuroClaimBench}: 489 executable neuroimaging claims that can be tested directly on the available imaging data. Each claim, analysis plan, and data partition is fixed before evaluation. Claims are scored against literature-grounded labels and 150 designed controls that violate individual methodological requirements. On NeuroClaimBench, CONFIRM falsely supports at most 3\% of
designed controls, compared with 23--38\% for the baselines, while recovering up
to 61\% of literature-supported claims.

Our contributions are:
\begin{itemize}
  \item \textbf{The reporting gap in agentic science.} We show that LLM judges and fixed heuristics accept 23--38\% of designed controls, which deliberately violate methodological requirements. They accept site-confounded findings even when shown evidence of the confound.
  \item \textbf{CONFIRM, an agentic framework for scientific reporting.} We combine LLM-based analysis planning and follow-up with deterministic execution and reporting gates. CONFIRM runs as a standalone agent or audits an existing agent.
  \item \textbf{NeuroClaimBench.} We provide frozen neuroimaging claims, literature reference labels, and designed negative controls for evaluating scientific reporting decisions.
\end{itemize}

\section{Related Work}
\label{sec:related}

\paragraph{Agentic neuroimaging and iterative refinement.}
LLMs can plan and execute end-to-end scientific workflows~\citep{boiko2023}. Neuroimaging agents have focused on pipeline reproducibility~\citep{neuroclaw2026}, preprocessing and disease prediction~\citep{neuroagent2026}, predictive accuracy~\citep{han2026}, or robustness against hallucinated inputs~\citep{neura2026}, while VERITAS uses fixed rules to assign evidence labels to hypotheses derived from images~\citep{veritas2026}. None of these systems determines whether a neuroimaging claim warrants scientific reporting, so Section~\ref{sec:ncb_v21_results} evaluates adaptations of the two closest frameworks. In general agent architectures, Self-Refine repeatedly critiques and revises its generated responses~\citep{madaan2023selfrefine}, but it neither diagnoses specific methodological failures nor delegates reporting decisions to deterministic code.

\paragraph{LLM judges.}
LLMs are frequently used as evaluators, but their judgments show position bias (sensitivity to response order), verbosity bias (favoring longer responses), and self-enhancement bias (favoring their own outputs)~\citep{zheng2023judging}. Scientific agents increasingly rely on LLM critics to decide what to report~\citep{neuroclaw2026}. CONFIRM assigns this reporting decision to code: the model proposes analysis plans and follow-up candidates, while deterministic gates decide which findings may be reported.

\paragraph{Governance layers and agent benchmarks.}
POPPER formulates sequential falsification experiments to evaluate free-form hypotheses with Type-I error control~\citep{huang2025popper}. CONFIRM fixes an observational analysis plan before computation and uses pre-specified criteria to decide which findings may be reported. It runs on its own or audits an existing agent. EviBound requires machine-checkable artifacts before an agent reports completed work~\citep{chen2025evibound} to verify task execution. ScienceAgentBench scores program code against references~\citep{chen2025scienceagentbench}. NeuroClaimBench evaluates which scientific claims the completed analyses support.

\paragraph{Statistical safeguards for neuroimaging.}
Techniques such as false discovery rate control~\citep{benjamini1995}, multiverse analysis~\citep{steegen2016,carp2012}, and ComBat harmonization~\citep{johnson2007,fortin2018} each target specific threats to validity. CONFIRM combines these safeguards into an automated reporting rule for scientific agents.
\section{CONFIRM: Frozen Claim Governance}
\label{sec:method}

\begin{figure*}[t]
\centering
\includegraphics[width=\textwidth]{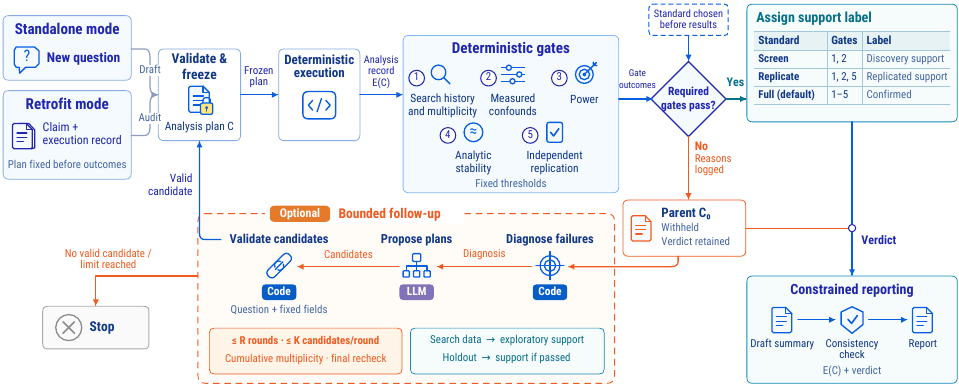}
\caption{\textbf{CONFIRM separates running an analysis from deciding whether its finding may be reported.} The LLM turns a question into an analysis plan that is frozen before any statistic is computed. Code runs the analysis, and five gates check the result. The reporting standard, chosen before results are seen, sets which gates must pass, and only Full can report a claim as \texttt{confirmed}. If a required gate fails, the claim is withheld, and the LLM may propose revised analyses that keep the original question. These are tested first on the discovery and replication cohorts and then on holdout data.}
\label{fig:architecture}
\end{figure*}

CONFIRM is a framework that determines whether an autonomous analysis agent may report a statistical claim as supported. To eliminate confirmation bias and selective reporting, CONFIRM separates three stages: specifying a claim, computing its results with deterministic code, and deciding whether to report it (Figure~\ref{fig:architecture}).

The framework operates in two modes. In \emph{standalone mode}, it governs the agent during execution and evaluates claims before any final report can be generated. In \emph{retrofit mode}, it audits an existing agent's claim and record of the analysis after execution. The original analysis plan must have been fixed and timestamped before results were observed. Both modes apply the same analysis plan format, verification gates, and reporting standards.

\subsection{Frozen Analysis Plans and Deterministic Execution}
\label{sec:contract}

Before fitting any statistical models, the agent translates a natural-language research question into a formal, machine-readable \emph{pre-specified analysis plan}:
\begin{equation}
C = (\theta, \mathcal{D}_{\mathrm{disc}}, \mathcal{D}_{\mathrm{rep}},
\mathbf{Z}, M_0, \mathcal{G}),
\label{eq:contract}
\end{equation}
where $\theta$ defines the target estimand, including the primary predictor or group contrast, the outcome variable, the hypothesized effect direction, and the claim scope (a scalar single-region measure vs.\ a brain-wide spatial pattern); $\mathcal{D}_{\mathrm{disc}}$ and $\mathcal{D}_{\mathrm{rep}}$ designate the discovery and independent replication cohorts; $\mathbf{Z}$ specifies the set of initial covariates used for adjustment; $M_0$ denotes the initial declared number of tests; and $\mathcal{G}$ sets the deterministic gate thresholds (Figure~\ref{fig:contract} shows an example).

\begin{figure}[t]
\centering
\begingroup
\definecolor{planblue}{HTML}{245980}
\definecolor{planrule}{HTML}{C9D9E6}
\definecolor{planfill}{HTML}{F5F9FC}
\setlength{\fboxsep}{9pt}
\newcommand{\planfield}[1]{\texttt{#1}}
\newcommand{\planhead}[1]{{\color{planblue}\sffamily\bfseries #1}}
\newcommand{\plangate}[2]{\planhead{#1}\par\vspace{1pt}{\footnotesize #2}\par\vspace{6pt}}
\noindent\colorbox{planfill}{%
\begin{minipage}{\dimexpr\linewidth-2\fboxsep\relax}
\small
\begin{minipage}[t]{0.49\linewidth}
\vspace{0pt}
\planhead{Claim and data}\par\vspace{3pt}
{\color{planrule}\hrule height 0.5pt}\vspace{6pt}
\renewcommand{\arraystretch}{1.18}
\setlength{\tabcolsep}{0pt}
\begin{tabular}{@{}>{\raggedright\arraybackslash}p{0.40\linewidth}
                  >{\raggedright\arraybackslash}p{0.60\linewidth}@{}}
Estimand type & \planfield{group\_diff} \\
Outcome & \planfield{smri\_hippocampus} \\
Group contrast & \planfield{dx}: Dementia (case),\newline CN (control) \\
Direction & Negative \\
Covariates & \planfield{age}, \planfield{sex}, \planfield{eTIV},\newline \planfield{site}, \planfield{scanner} \\
Discovery cohort & ADNI \\
Replication cohort & OASIS3 \\
Search family & 1 declared comparison \\
\end{tabular}
\end{minipage}\hfill
\begin{minipage}[t]{0.46\linewidth}
\vspace{0pt}
\planhead{Fixed verification gates}\par\vspace{3pt}
{\color{planrule}\hrule height 0.5pt}\vspace{6pt}
\plangate{1\quad Search history and multiplicity}{%
\planfield{method: bonferroni}; $\alpha=0.05$}
\plangate{2\quad Measured confounds}{%
Required covariates: \planfield{site}, \planfield{scanner}}
\plangate{3\quad Power}{%
\planfield{min\_power}: $0.80$}
\plangate{4\quad Analytic stability (\texttt{multiverse})}{%
\planfield{min\_consistent}: $0.60$}
\planhead{5\quad Independent replication}\par\vspace{1pt}
{\footnotesize Same sign: \planfield{true}; $\alpha=0.05$;\newline
\planfield{harmonize: combat}}
\end{minipage}
\par\vspace{7pt}
{\color{planrule}\hrule height 0.5pt}\vspace{5pt}
{\footnotesize\planhead{Allowed report labels}\quad
\planfield{confirmed}\enspace\textperiodcentered\enspace
\planfield{fragile}\enspace\textperiodcentered\enspace
\planfield{under\_powered}\enspace\textperiodcentered\enspace
\planfield{non\_replicated}}
\end{minipage}%
}
\endgroup

\caption{\textbf{An abbreviated analysis plan for hippocampal atrophy in
Alzheimer's disease.} The frozen analysis plan fixes the effect, covariates,
cohorts, search family, and gate thresholds, and deterministic code supplies
every numeric result after the plan is frozen. The \texttt{multiverse} field
implements the analytic-stability gate.}
\label{fig:contract}
\end{figure}

The analysis plan $C$ is checked against a schema that specifies its required fields and format. Before any outcome data are read, the plan is fixed and assigned a cryptographic hash, an identifier computed from its contents. The plan is then locked, so the agent cannot alter the outcome variable, covariates, multiplicity penalties, or replication criteria after inspecting results.

Once the plan is frozen, deterministic scripts fit the models and generate an analysis record $E(C)$ containing effect estimates, multiverse analyses (repeated fits under pre-specified alternative analysis choices), power calculations, replication statistics, and checks of the study design. The LLM agent receives $E(C)$ solely to draft natural-language summaries. An automated consistency check validates all numbers and labels in the draft against $E(C)$, discarding any value that differs from the analysis record.

\subsection{Deterministic Verification Gates and Reporting Standards}
\label{sec:gates}

CONFIRM subjects every claim to five deterministic checks, termed \emph{gates}. Each gate targets a specific failure mode in empirical neuroimaging (Table~\ref{tab:gate_overview}). Formally, the gates evaluate the analysis plan and execution record, mapping $(C, E)$ to a vector of pass/fail results, $\mathbf{g}(C,E) \in \{0, 1\}^5$. All gate thresholds $\mathcal{G}$ are fixed in $C$ before execution and cannot be changed after results are observed.

\begin{table*}[t]
\centering
\small
\setlength{\tabcolsep}{4pt}
\caption{\textbf{Each gate catches one known way a neuroimaging finding can go wrong.}
The rightmost column lists the designed controls used to benchmark each gate (Section~\ref{sec:benchmark}). Random-label controls lack true signal and trigger multiple gates. Exact mathematical definitions for scalar and brain-wide claims appear in Appendix Table~\ref{tab:gate_thresholds}.}
\label{tab:gate_overview}
\begin{tabular}{@{}>{\raggedright\arraybackslash}p{0.23\textwidth}
                >{\raggedright\arraybackslash}p{0.15\textwidth}
                >{\raggedright\arraybackslash}p{0.39\textwidth}
                >{\raggedright\arraybackslash}p{0.15\textwidth}@{}}
\toprule
What can go wrong & Gate & Passes when (default) & Designed control \\
\midrule
Many tests were tried and only the best one is reported & Search history and multiplicity & The effect stays significant after correcting for every test tried ($\alpha=0.05$). & Broad-search \\
\addlinespace
A site or scanner difference creates an apparent effect & Measured confounds & No measured site, scanner, field-strength, or software variable differs between the groups while being left out of the analysis. & Site-confounded \\
\addlinespace
The sample is too small to trust the result & Power & The planned design can detect a modest reference effect, by default $d=0.3$, with at least 80\% power. & Underpowered \\
\addlinespace
The result depends on arbitrary analysis choices & Analytic stability & At least 60\% of reasonable alternative analyses give a consistent result~\citep{steegen2016}. & -- \\
\addlinespace
The effect does not appear again in new data & Independent replication & The effect has the same direction and $p<0.05$ in every replication dataset. & Nonreplicating \\
\bottomrule
\end{tabular}
\end{table*}

The \textbf{search history and multiplicity} gate accounts for undeclared tests by correcting the claim for the effective number of tests,
\begin{equation}
M_{\mathrm{eff}} = \max\left\{M_0,\, M_{\mathrm{history}},\, M_{\mathrm{observed}}\right\},
\end{equation}
where $M_0$ is the declared count, $M_{\mathrm{history}}$ is the count logged in the agent's interaction history, and $M_{\mathrm{observed}}$ is the number of features actually fitted (Appendix Table~\ref{tab:gate_thresholds}). For \textbf{statistical power}, the gate computes prospective power: the probability of detecting a reference effect size $\delta_{\mathrm{ref}}$ chosen before results are observed. When literature values are unavailable, CONFIRM defaults to $\delta_{\mathrm{ref}} = 0.3$ (Cohen's $d$). No benchmark claim in our evaluation had a reference effect size established in the literature, so all benchmark power gates use this default, and Table~\ref{tab:robustness_compact} reports sensitivity to alternative values.

A claim is designated \texttt{confirmed} if and only if its frozen analysis passes all five gates ($\mathbf{g} = \mathbf{1}$) on the designated discovery and replication cohorts. This label denotes support under the Full reporting standard. The claim is withheld if any gate fails or required diagnostic metadata are missing. Withheld claims receive an explanatory diagnostic label according to the following priority order: \texttt{fragile} (multiplicity, confound, or stability failure), \texttt{under\_powered} (power failure), or \texttt{non\_replicated} (replication failure).

Because gate checks are deterministic and computed once, CONFIRM supports three reporting standards, each adding requirements to the previous one. CONFIRM-Screen requires only the multiplicity and confound gates and reports \emph{discovery support}. CONFIRM-Replicate adds the independent replication gate and reports \emph{replicated support}. CONFIRM-Full, the default, enforces all five gates and is the only standard that can report a claim as \texttt{confirmed}.

\subsection{Bounded, Failure-Directed Follow-Up Search}
\label{sec:followup-search}

When a parent analysis plan $C_0$ fails, CONFIRM uses the failed gates to guide a bounded search for follow-up analyses. The parent retains its withheld verdict, and each follow-up receives a separate evaluation.

A deterministic diagnostic routine first identifies the failed gates and the fields that must stay fixed. The search then runs for at most $R$ rounds, and each round $r \in \{1, \dots, R\}$ has three steps. First, the LLM agent receives $C_0$, the failure diagnostic, the available dataset variables, and the history of failed candidates from earlier rounds, and it proposes up to $K$ candidate analysis plans $\{C_{r,1}, \dots, C_{r,K}\}$. Second, a strict validator rejects any candidate that alters the core scientific question, the primary predictor or contrast, the hypothesized direction, the underlying datasets, or the gate thresholds, or that removes a previously included covariate. Third, surviving candidates are executed only on the parent's discovery and replication cohorts (the \emph{search data}). To penalize iterative exploration, candidate $k$ in round $r$ is evaluated against a cumulative multiplicity penalty
\begin{equation}
M_{\mathrm{eff}}(C_{r,k}, r) = \max\left\{M_{\mathrm{declared}}(C_{r,k}),\, M_0 + H_r\right\},
\end{equation}
where $H_r$ is the cumulative count of unique hypotheses evaluated through round $r$. Any candidate that provisionally passes is re-checked against the final hypothesis count $H_R$ at the end of the search.

Candidates surviving this follow-up protocol receive \emph{exploratory support} on the search data. Support for the frozen follow-up plan is assessed on additional data: designated holdout and external cohorts are kept out of candidate generation and follow-up search. They are used only after the candidate plan and its final count of tests are fixed. Appendix~\ref{sec:appendix_method_details} specifies how failed gates guide revisions, when search stops, and how candidates are validated.
\section{NeuroClaimBench: A Claim-Level Benchmark}
\label{sec:benchmark}

\ifvenueaamas\begin{figure*}[t]
\centering
\includegraphics[width=\textwidth]{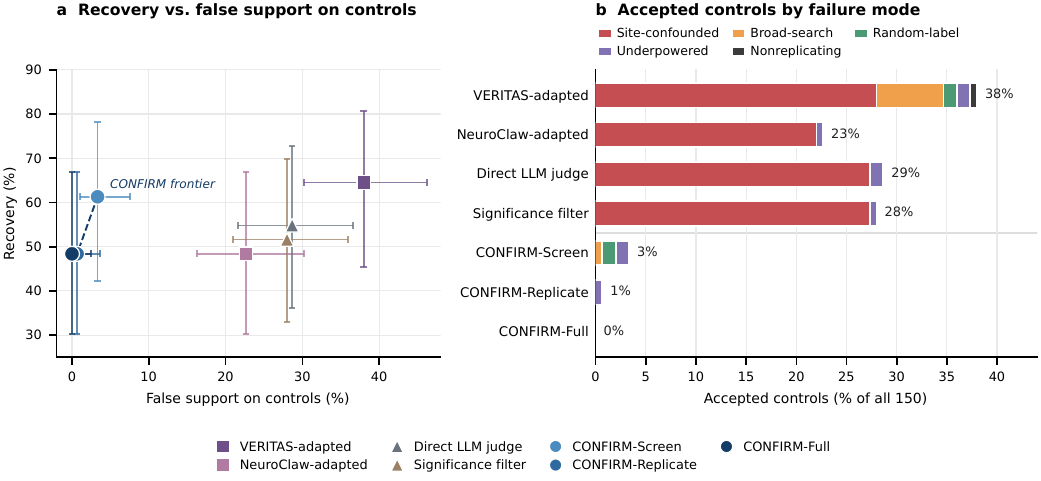}
\caption{\textbf{CONFIRM establishes a Pareto-optimal reporting frontier.}
\textbf{a}, Recovery on literature-supported claims in the common scalar set (higher is better) versus false support on designed negative controls (lower is better). The dashed curve connects CONFIRM's three nested reporting standards (Screen, Replicate, Full). Error bars indicate 95\% confidence intervals.
\textbf{b}, Accepted controls by failure mode. Each segment is a percentage of all 150 controls (42 site-confounded, 42 broad-search, 42 random-label, 21 underpowered, and 3 nonreplicating).}
\label{fig:confirm_frontier}
\end{figure*}
\fi

Existing benchmarks for AI analysis agents typically evaluate \emph{code execution}: whether a model writes code that runs to completion. NeuroClaimBench evaluates \emph{reporting decisions}: whether a system agrees with literature-grounded references and withholds claims that violate the specified methodological requirements.

Benchmark cases are defined independently of the outputs of the systems being evaluated. Each benchmark item pairs a natural-language research question with a frozen, executable analysis plan, its designated data partitions, and an adjudicated reference label where one exists.

\subsection{Benchmark Construction and Evidence Partitions}
\label{sec:benchmark_construction}

We assessed each candidate item before execution, without access to its results. We checked that the research question matched the analysis plan and datasets, and that the specified estimand was statistically identifiable, meaning that it could be uniquely determined from the planned model and available data.

Out of 544 candidate items, 55 ambiguous or non-executable items were excluded. The remaining 489 executable claims comprise 279 scientific claims drawn from neuroimaging cohorts, 60 external-transfer claims that evaluate generalization to other imaging consortia, and 150 designed negative controls, which are claims with deliberate methodological flaws and a known negative label.
The counts for each category appear in Appendix Table~\ref{tab:ncb_v21_composition}. The benchmark spans structural MRI (sMRI), resting-state functional connectivity (rs-fMRI), and positron emission tomography (PET), covering healthy populations, attention-deficit/hyperactivity disorder (ADHD), autism spectrum disorder (ASD), Alzheimer's disease/aging, and psychosis. Each analysis plan designates discovery, replication, and holdout partitions with no participants shared between them.

\paragraph{Designed negative controls.}
New scientific claims often lack a known reference answer. We therefore constructed 150 designed negative controls whose methodological flaws require withholding the claim. Each follows a standard neuroimaging study design with one deliberate flaw that violates a reporting requirement.

The controls represent five distinct failure modes:
\begin{enumerate}[leftmargin=*,noitemsep,topsep=2pt]
    \item \textbf{Site-confounded:} Comparison groups are sampled from different acquisition sites in unbalanced ratios, mixing the apparent group effect with scanner differences.
    \item \textbf{Broad-search:} Many tests are run, and a significant result is selected for reporting without correction for the number of tests.
    \item \textbf{Random-label:} Group diagnostic labels are randomly permuted, ensuring that any detected effect represents sample noise.
    \item \textbf{Underpowered:} Sample sizes are severely restricted, so the design lacks 80\% power to detect the reference effect.
    \item \textbf{Nonreplicating:} An effect observed in the discovery cohort fails to replicate ($p \ge 0.05$ or the opposite sign) in an independent cohort.
\end{enumerate}

\subsection{Literature Reference Labels and Adjudication}
\label{sec:benchmark_labels}

Literature reference labels were established for 96 scientific and external-transfer claims through agreement among LLM assessors and expert review. Two independent LLMs (GPT-5.5 and Gemini 3.5 Flash) evaluated fixed sets of PubMed evidence for each contrast. An independent arbitrator (Claude Opus 4.8) cast the deciding vote, and neuroimaging experts reviewed and finalized all resulting labels.
This process yielded 47 \emph{strict} labels (requiring agreement between the arbitrator and at least one assessor, followed by expert consensus) and 49 \emph{provisional} labels with weaker agreement among assessors (Appendix~\ref{sec:appendix_reference_construction}). The 150 designed controls and 22 random-label controls in the external-transfer split have a \emph{withhold} reference label. The remaining 221 executable claims lack direct literature evidence for their exact contrast and are excluded when calculating the metrics.

\subsection{Evaluation Subsets and Performance Metrics}
\label{sec:scoring_sets}

Evaluation uses three subsets. The \textbf{literature set} ($N = 70$) contains scientific claims with adjudicated literature labels, 51 supported and 19 unsupported, covering both scalar brain measures (e.g., hippocampal volume) and brain-wide patterns. Because baseline methods (except direct LLM judges) have no decision rule for claims about patterns across many brain regions, head-to-head comparisons use the \textbf{common scalar set} ($N = 46$) of single-measure claims, with 31 supported and 15 unsupported. The \textbf{control set} ($N = 150$) contains all designed negative controls, all of which should be rejected.
If a baseline method fails to run or returns no verdict for an item, the item counts as withheld.

System performance is summarized with two metrics. \textbf{Recovery} ($\uparrow$) is the proportion of literature-supported claims that a method accepts. \textbf{False support} ($\downarrow$) is the proportion of claims with a negative reference label that a method accepts, reported separately for designed controls and literature-unsupported claims. 
\section{Experiments}
\label{sec:experiments}

\ifvenueaamas\begin{figure*}[tb]
\centering
\includegraphics[width=0.85\textwidth]{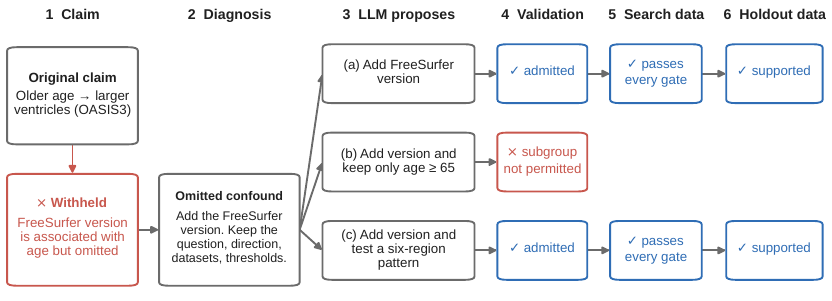}
\caption{\textbf{CONFIRM's diagnosis leads the LLM to a revision supported on holdout data.}
CONFIRM withholds a claim because the FreeSurfer version, which is associated with age, was omitted. The LLM's two valid revisions add the version, pass every gate on the search data, and are supported on holdout data. Blue boxes mark steps that pass, and red boxes mark steps that fail.}
\label{fig:worked_example}
\end{figure*}
\fi

We ask three questions. Does CONFIRM eliminate methodologically flawed findings without sacrificing literature-supported claims (Section~\ref{sec:ncb_v21_results})? Do CONFIRM's deterministic failure diagnoses guide agents toward revisions supported on search and holdout data (Section~\ref{sec:feedback_results})? Can CONFIRM be integrated into existing autonomous agents to audit and improve their reliability (Section~\ref{sec:neuroclaw_feedback})? Implementation details, parameter settings, and additional robustness checks appear in Appendix~\ref{sec:appendix_experiment_details}.

\subsection{CONFIRM Rejects Designed Controls While Recovering Literature-Supported Claims}
\label{sec:ncb_v21_results}

We compare CONFIRM against four reporting policies:
\emph{VERITAS-adapted} (fixed heuristic)~\citep{veritas2026},
\emph{NeuroClaw-adapted} (LLM critic panel)~\citep{neuroclaw2026},
a \emph{direct LLM judge} (prompted on summary statistics and metadata), and
a \emph{significance filter} (reporting any directional $p < 0.05$ across datasets).
We adapt VERITAS and NeuroClaw to evaluate frozen claims using their original decision rules and persona prompts, respectively. VERITAS-adapted reimplements VERITAS's fixed rule for assigning evidence labels with its default smallest effect of interest ($d=0.5$, $r=0.3$), $\alpha=0.05$, and its power formula. NeuroClaw-adapted uses GPT-5.5 with NeuroClaw's original prompts for three roles: biostatistician, clinical neuroscientist, and methodology expert. It applies NeuroClaw's panel rule, including the methodology expert's veto, and reports a claim only when all three roles support it. No threshold, prompt, or aggregation rule was tuned on NeuroClaimBench (Appendix~\ref{sec:appendix_comparison_protocol}). All methods receive the same evidence as CONFIRM: the discovery and replication effects, results from the alternative analyses, power, the declared search family, and each measured site or scanner variable's association with the predictor. No method sees a reference label, gate state, or CONFIRM verdict.

\ifvenueaamas\begin{figure*}[t]
\centering
\includegraphics[width=0.9\textwidth]{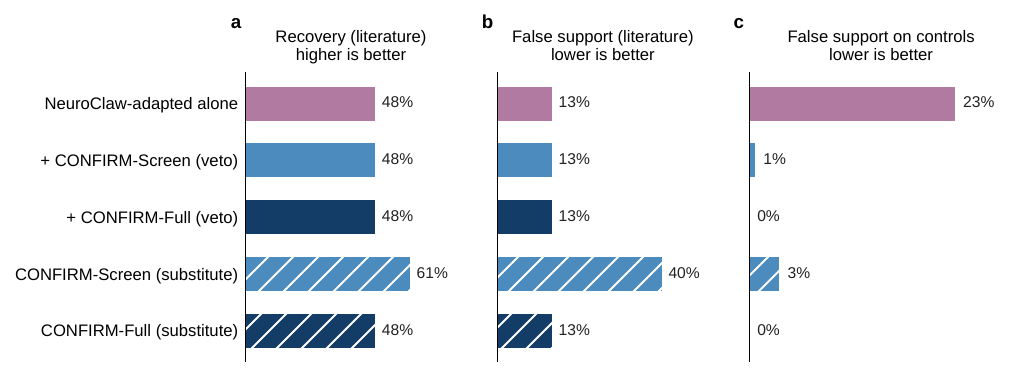}
\caption{\textbf{Adding CONFIRM as a retrofit layer to NeuroClaw-adapted reduces an existing agent's false support without losing its findings.} Solid bars depict a filter where a claim is reported only if both NeuroClaw-adapted and CONFIRM accept it. Hatched bars replace the agent's reporting decision with CONFIRM's decision. Metrics are evaluated on the common scalar set: \textbf{a}, Recovery on literature-supported claims (higher is better); \textbf{b}, False support on literature-unsupported claims (lower is better); and \textbf{c}, False support on designed negative controls (lower is better).}
\label{fig:retrofit_composition}
\end{figure*}
\fi

\ifvenueaamas\else\begin{figure*}[t]
\centering
\includegraphics[width=\textwidth]{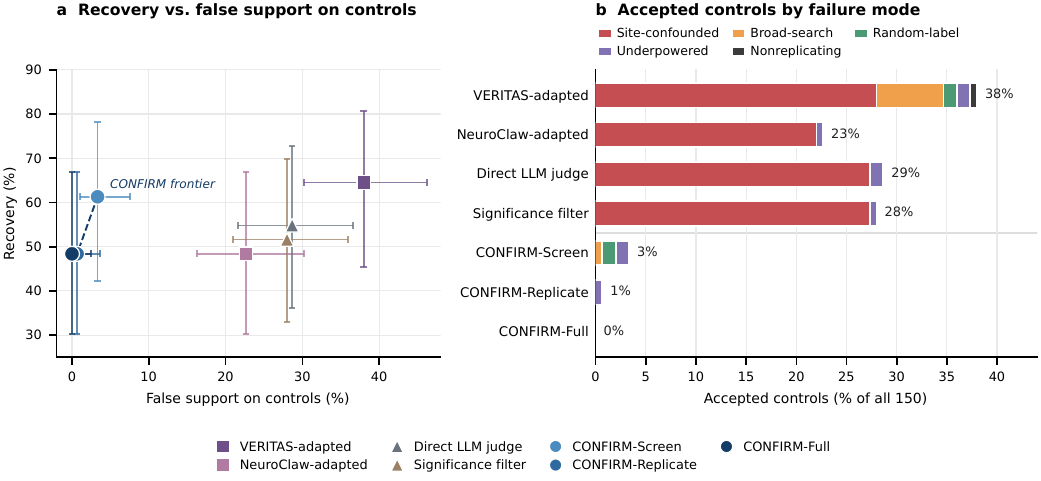}
\caption{\textbf{CONFIRM establishes a Pareto-optimal reporting frontier.}
\textbf{a}, Recovery on literature-supported claims in the common scalar set (higher is better) versus false support on designed negative controls (lower is better). The dashed curve connects CONFIRM's three nested reporting standards (Screen, Replicate, Full). Error bars indicate 95\% confidence intervals.
\textbf{b}, Accepted controls by failure mode. Each segment is a percentage of all 150 controls (42 site-confounded, 42 broad-search, 42 random-label, 21 underpowered, and 3 nonreplicating).}
\label{fig:confirm_frontier}
\end{figure*}
\fi

Baselines accept 23--38\% of designed negative controls. CONFIRM accepts at most 3\%, and CONFIRM-Full accepts none (Figure~\ref{fig:confirm_frontier}a and Appendix Table~\ref{tab:claim_evaluation_common_scalar}). On the common scalar set (Section~\ref{sec:scoring_sets}), CONFIRM's recovery on literature-supported claims matches or approaches that of the baselines, establishing a Pareto-optimal frontier for the trade-off between recovery and false support. On the full literature set, CONFIRM-Full withholds 59\% of literature-supported claims, and 87\% of these fail to replicate in the independent cohort. Most also fail the multiplicity or stability gate. These effects are reported in the literature but do not hold in these datasets under a pre-specified analysis (Appendix Table~\ref{tab:withheld_gates}).

Figure~\ref{fig:confirm_frontier}b separates accepted controls by failure type. Baselines fail primarily on site confounding, accepting 79--100\% of site-confounded controls. LLM-based judges accept these controls even though they are given the site's association with the predictor, while rule-based baselines ignore confounds entirely. CONFIRM's confound gate rejects all of them at every standard. Stricter standards filter the remaining controls through replication, prospective power, and stability gates.

These findings hold when evaluating brain-wide spatial patterns and strict consensus labels (Appendices~\ref{sec:appendix_benchmark_details} and~\ref{sec:appendix_experiment_details}). When claims about patterns across many brain regions are added, CONFIRM-Screen accepts 55\% of literature-supported claims against 41\% for the direct LLM judge, and CONFIRM-Full matches the judge's 41\% while accepting half as many unsupported claims, 11\% against 21\% (Appendix Figure~\ref{fig:confirm_policy_comparison}). Changing any single gate threshold never causes CONFIRM-Full to accept a control (Table~\ref{tab:robustness_compact}). An ablation removes one gate at a time to measure its contribution. Removing the measured-confound gate raises CONFIRM-Full's false support on designed controls by 24 percentage points, and removing the replication gate raises its false support on literature-unsupported claims by 26 points (Appendix Table~\ref{tab:confirm_robustness}).

\begin{table}[!tb]
\centering
\small
\setlength{\tabcolsep}{4pt}
\caption{\textbf{CONFIRM-Full rejects all designed controls across the tested gate thresholds.} Each row changes one
setting and keeps all others at their defaults. Rec.\ is recovery on 51
literature-supported claims, FS is false support on 19 literature-unsupported
claims, and Ctrl.\ is false support on 150 designed controls. Numbers shown are percentages (\%). Brain-wide replication settings change the pattern-correlation, replicated-region-fraction, and Dice thresholds, which set how closely a brain-wide pattern must match across cohorts to count as replicated.}
\label{tab:robustness_compact}
\begin{tabular}{@{}lrrr@{}}
\toprule
Setting & Rec.\ $\uparrow$ & FS $\downarrow$ & Ctrl.\ $\downarrow$ \\
\midrule
Defaults & 41 & 11 & 0 \\
\addlinespace
\multicolumn{4}{@{}l}{\textit{Gate thresholds}} \\
Stability $\rho \in \{0.5, 0.7, 0.8\}$ & 41 & 11 & 0 \\
Minimum power $\in \{0.7, 0.9\}$ & 41 & 11 & 0 \\
Confound $\alpha \in \{0.01, 0.10\}$ & 35--41 & 11 & 0 \\
Replication $\alpha \in \{0.01, 0.10\}$ & 37--41 & 5--11 & 0 \\
Brain-wide replication (3 settings) & 41 & 11--16 & 0 \\
\addlinespace
\multicolumn{4}{@{}l}{\textit{Power reference effect}} \\
$d \in \{0.2, 0.4, 0.5\}$ & 41 & 11 & 0--1 \\
$d = 0.1$ & 14 & 0 & 0 \\
\bottomrule
\end{tabular}
\end{table}

\ifvenueaamas\else\begin{table}[!tb]
\centering
\small
\setlength{\tabcolsep}{6pt}
\renewcommand{\arraystretch}{1.12}
\caption{\textbf{CONFIRM diagnostics yield more supported revisions with fewer LLM calls.} GPT-5.5 revises the same 215 withheld claims for three rounds, with up to five candidates per round. Methods share the validator and multiplicity correction and differ only in failure feedback.}
\label{tab:feedback_control_main}
\begin{tabular}{@{}lrrrrr@{}}
\toprule
Method & LLM calls & Candidates & Supported & Parents & Holdout \\
 & $\downarrow$ & evaluated & on search $\uparrow$ & (\%) $\uparrow$ & (\%) $\uparrow$ \\
\midrule
Failure-blind retry & 765 & 3,073 & 49 & 9 & 7 \\
Self-Refine & 1,301 & 2,161 & 54 & 11 & 9 \\
CONFIRM diagnosis & 653 & 2,971 & 70 & 11 & 17 \\
\bottomrule
\end{tabular}
\par\vspace{4pt}
\begin{minipage}{0.94\textwidth}\footnotesize\raggedright
\textit{Supported}: candidates passing all gates on the search data. \textit{Parents}: withheld claims with at least one supported revision. \textit{Holdout}: the percentage of candidates tested on holdout data that are supported there.
\end{minipage}
\end{table}
\fi

\subsection{Deterministic Failure Diagnoses Guide Effective Follow-Up}
\label{sec:end_to_end}
\label{sec:feedback_results}

Run autonomously, CONFIRM formulated and executed 289 analysis plans. CONFIRM-Full withheld 74\% of them, citing the exact failed gates. When evaluated on holdout partitions, 25\% of initially confirmed claims passed the Full reporting standard again, compared with 3\% of withheld claims. The replication gate alone passed for 63\% and 15\%, respectively (Appendix~\ref{sec:appendix_heldout_verdicts}). The 215 withheld claims are the starting point for the follow-up search below.

\ifvenueaamas\begin{table}[!tb]
\centering
\small
\setlength{\tabcolsep}{6pt}
\renewcommand{\arraystretch}{1.12}
\caption{\textbf{CONFIRM diagnostics yield more supported revisions with fewer LLM calls.} GPT-5.5 revises the same 215 withheld claims for three rounds, with up to five candidates per round. Methods share the validator and multiplicity correction and differ only in failure feedback.}
\label{tab:feedback_control_main}
\begin{tabular}{@{}lrrrrr@{}}
\toprule
Method & LLM calls & Candidates & Supported & Parents & Holdout \\
 & $\downarrow$ & evaluated & on search $\uparrow$ & (\%) $\uparrow$ & (\%) $\uparrow$ \\
\midrule
Failure-blind retry & 765 & 3,073 & 49 & 9 & 7 \\
Self-Refine & 1,301 & 2,161 & 54 & 11 & 9 \\
CONFIRM diagnosis & 653 & 2,971 & 70 & 11 & 17 \\
\bottomrule
\end{tabular}
\par\vspace{4pt}
\begin{minipage}{0.94\textwidth}\footnotesize\raggedright
\textit{Supported}: candidates passing all gates on the search data. \textit{Parents}: withheld claims with at least one supported revision. \textit{Holdout}: the percentage of candidates tested on holdout data that are supported there.
\end{minipage}
\end{table}
\fi

CONFIRM uses failed-gate diagnoses to guide follow-up analysis. We gave each withheld claim to GPT-5.5 and asked it to propose revised analysis plans within a fixed search budget ($R=3$ rounds, $K=5$ candidates per round), enforcing cumulative multiplicity corrections and schema validation. We compared three kinds of feedback. With \textbf{failure-blind retry}, the LLM receives only a binary notification of failure. With \textbf{Self-Refine}~\citep{madaan2023selfrefine}, the LLM is prompted to critique and revise its failed analysis on its own. With \textbf{CONFIRM diagnosis}, the LLM receives deterministic diagnostic output identifying the failed gates and the fields it must keep unchanged.

Across three independent evaluation runs (Table~\ref{tab:feedback_control_main} reports a representative run), CONFIRM diagnosis produced more revisions supported on the search data (70 supported revisions vs.\ $\le 54$ for alternatives). It also used fewer total LLM calls and reached approximately twice the holdout support rate of failure-blind retry or Self-Refine. For designed controls, a high-budget follow-up search ($R=10$, $K=10$) evaluated 10,716 candidates, and none passed even provisionally (Appendix Table~\ref{tab:feedback_control}).

\ifvenueaamas\else
\begin{figure*}[tb]
\centering
\includegraphics[width=0.85\textwidth]{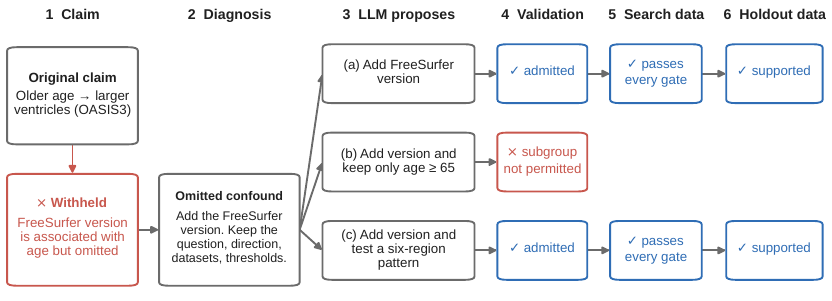}
\caption{\textbf{CONFIRM's diagnosis leads the LLM to a revision supported on holdout data.}
CONFIRM withholds a claim because the FreeSurfer version, which is associated with age, was omitted. The LLM's two valid revisions add the version, pass every gate on the search data, and are supported on holdout data. Blue boxes mark steps that pass, and red boxes mark steps that fail.}
\label{fig:worked_example}
\end{figure*}

\begin{figure*}[t]
\centering
\includegraphics[width=0.9\textwidth]{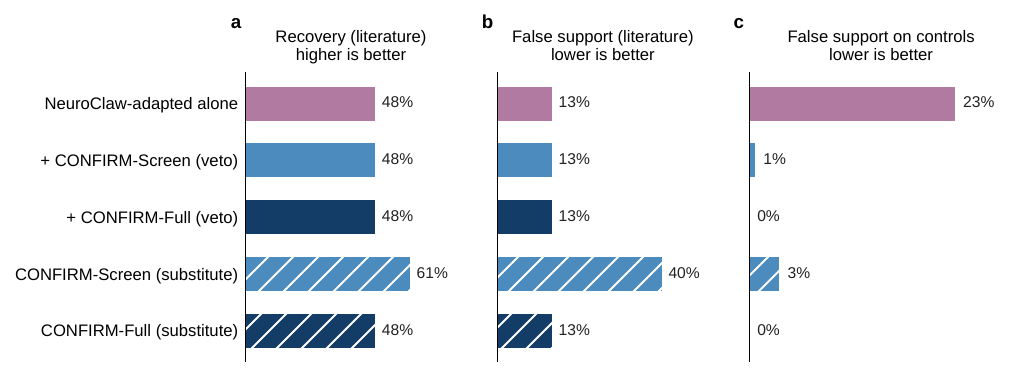}
\caption{\textbf{Adding CONFIRM as a retrofit layer to NeuroClaw-adapted reduces an existing agent's false support without losing its findings.} Solid bars depict a filter where a claim is reported only if both NeuroClaw-adapted and CONFIRM accept it. Hatched bars replace the agent's reporting decision with CONFIRM's decision. Metrics are evaluated on the common scalar set: \textbf{a}, Recovery on literature-supported claims (higher is better); \textbf{b}, False support on literature-unsupported claims (lower is better); and \textbf{c}, False support on designed negative controls (lower is better).}
\label{fig:retrofit_composition}
\end{figure*}

\begin{figure}[!tb]
\centering
\newcommand{\neuroclawfeedbackcaption}{\textbf{CONFIRM diagnoses improve an existing agent's revisions.} NeuroClaw-adapted drafts every claim, and the agent uses feedback to revise only the claims that CONFIRM withheld. Both panels are percentages of the 280 executed drafts. None of the three versions accepts a site-confounded control.}
\ifvenueaamas
\includegraphics[width=\smallfigwidth]{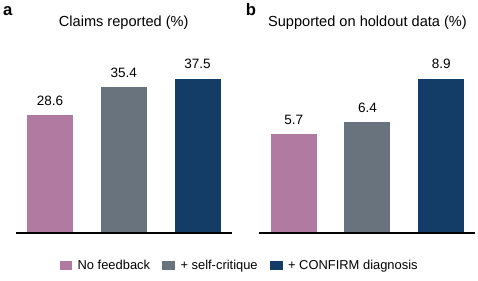}
\caption{\neuroclawfeedbackcaption}
\label{fig:neuroclaw_feedback}
\else
\begin{minipage}[c]{0.54\textwidth}
\centering
\includegraphics[width=\linewidth]{figures/fig_neuroclaw_feedback_column.pdf}
\end{minipage}\hfill
\begin{minipage}[c]{0.42\textwidth}
\caption{\neuroclawfeedbackcaption}
\label{fig:neuroclaw_feedback}
\end{minipage}
\fi
\end{figure}

\fi

Figure~\ref{fig:worked_example} traces one withheld claim through this process. CONFIRM withholds the claim that older age predicts larger ventricular volume in cognitively normal OASIS3 participants, because the FreeSurfer version used to process the scans is associated with age but omitted from the analysis. Given this diagnosis, the LLM proposes three revisions. The validator rejects one that restricts the sample to a subgroup outside the permitted set. The two valid revisions add the software version, pass every gate on the search data, and are supported on holdout data, where the original claim is not. Appendix~\ref{sec:appendix_worked_example} walks through each step.

\subsection{Retrofitting CONFIRM Onto Existing Autonomous Agents}
\label{sec:neuroclaw_feedback}

\ifvenueaamas\begin{figure}[!tb]
\centering
\newcommand{\neuroclawfeedbackcaption}{\textbf{CONFIRM diagnoses improve an existing agent's revisions.} NeuroClaw-adapted drafts every claim, and the agent uses feedback to revise only the claims that CONFIRM withheld. Both panels are percentages of the 280 executed drafts. None of the three versions accepts a site-confounded control.}
\ifvenueaamas
\includegraphics[width=\smallfigwidth]{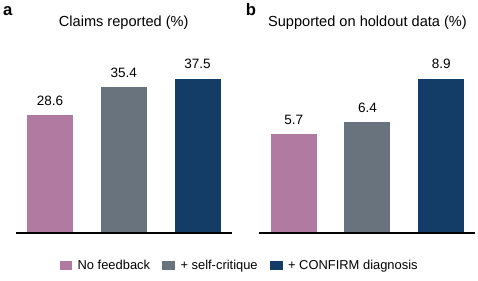}
\caption{\neuroclawfeedbackcaption}
\label{fig:neuroclaw_feedback}
\else
\begin{minipage}[c]{0.54\textwidth}
\centering
\includegraphics[width=\linewidth]{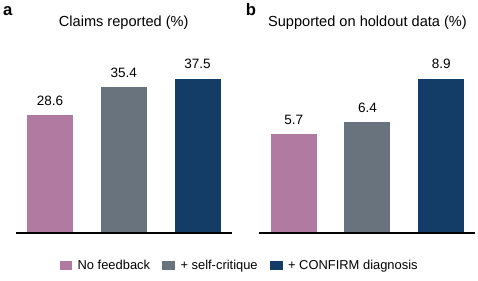}
\end{minipage}\hfill
\begin{minipage}[c]{0.42\textwidth}
\caption{\neuroclawfeedbackcaption}
\label{fig:neuroclaw_feedback}
\end{minipage}
\fi
\end{figure}
\fi

We also used CONFIRM to audit claims after NeuroClaw-adapted completed its analyses. NeuroClaw-adapted alone accepts 23\% of designed controls. Requiring its claims to pass CONFIRM-Full eliminates all false support on controls (0\%) while retaining every literature-supported claim the agent originally supported (Figure~\ref{fig:retrofit_composition}). CONFIRM's diagnoses also improve the agent's own revisions. When NeuroClaw-adapted revises its withheld claims, CONFIRM's diagnosis raises the share of drafts the agent reports from 28.6\% to 37.5\%, against 35.4\% with self-critique, while using fewer LLM calls. It also raises the share of drafts supported on holdout data from 5.7\% to 8.9\%, against 6.4\% with self-critique (Figure~\ref{fig:neuroclaw_feedback}, Appendix~\ref{sec:appendix_neuroclaw_feedback}).

\section{Discussion}
\label{sec:discussion}

\paragraph{Auditable reporting decisions.}
LLM judges apply available evidence inconsistently, while fixed heuristics omit relevant checks. The frozen analysis plan and execution record make the reporting criteria and their outcomes available for inspection. Users can trace a withheld claim to its failed checks, and agents can use those diagnoses to plan follow-up analyses. The same records support external audits across agents.

\paragraph{Reporting standards for exploration and confirmation.}
The reporting standards let researchers match requirements to the study's purpose. CONFIRM-Screen supports exploratory analysis and screening many claims, CONFIRM-Replicate supports multi-cohort studies with limited statistical power, and CONFIRM-Full suits confirmatory studies that must pass all five gates. The standard is selected before results are observed. Stored gate outcomes also let readers apply stricter standards without rerunning models.

\paragraph{Generalization across scientific fields.}
Applying CONFIRM in another field requires translating its methodological requirements into executable gates, such as checks for batch effects in genomics, selection bias in epidemiology, or invalid instrumental variables in econometrics. The same benchmark design can be used in those fields: construct negative controls that violate the relevant requirements and measure how often agents support them.
\section{Conclusion}
\label{sec:conclusion}

LLM agents that run neuroimaging analyses still leave the decision to report a finding to LLM judgment, and LLM judges endorse site-confounded findings even when the confound is shown to them. CONFIRM takes this decision away from the LLM. The LLM writes a pre-specified analysis plan, code runs it, and five deterministic gates decide whether the finding may be reported under a reporting standard chosen before results are seen. Every verdict records which gates passed and which failed, so a withheld claim can be traced to its cause.

On NeuroClaimBench, CONFIRM falsely supports at most 3\% of designed controls, compared with 23--38\% for the baselines, and CONFIRM-Full supports none. It recovers up to 61\% of literature-supported claims, and 87\% of the literature-supported claims that CONFIRM-Full withholds fail to replicate in the independent cohort. Its diagnoses tell the LLM which gate failed and which parts of the analysis plan a revision must preserve. With them, the LLM finds more revisions supported on the search data than with failure-blind retry or Self-Refine, reaches about twice their holdout support rate, and uses fewer LLM calls. Used to audit an existing agent, CONFIRM lowers false support on designed controls from 23\% to 0\% and keeps every literature-supported claim the agent supported.

These results show that an agent can propose and run analyses freely, while criteria fixed before results are observed determine which findings it may report. The same separation carries over to other fields wherever methodological requirements can be written as executable checks, and designed controls like those in NeuroClaimBench measure how often an agent reports findings that violate them.

\FloatBarrier

\bibliography{references}
\bibliographystyle{tmlr}

\clearpage
\appendix
\section*{Appendix Overview}
\label{sec:appendix_roadmap}

Appendix~\ref{sec:appendix_method_details} specifies the gate implementations,
verdict rules, and follow-up search policy.
Appendix~\ref{sec:appendix_benchmark_details} describes how NeuroClaimBench and
its reference labels were constructed.
Appendix~\ref{sec:appendix_experiment_details} details the comparison protocol,
robustness analyses, and agent-integration experiments, and walks through one
withheld claim step by step. Appendix~\ref{sec:limitations} states the scope of
the study.

\section{Gate and Follow-Up Implementation Details}
\label{sec:appendix_method_details}

\subsection{Gate Implementation}

\begin{table*}[t]
\centering
\small
\setlength{\tabcolsep}{6pt}
\caption{\textbf{Exact gate decision rules and default thresholds.} This table
is the formal counterpart of Table~\ref{tab:gate_overview}. Analysis plans may
specify stricter thresholds.}
\label{tab:gate_thresholds}
\begin{tabular}{@{}>{\raggedright\arraybackslash}p{0.17\textwidth}
                >{\raggedright\arraybackslash}p{0.56\textwidth}
                >{\raggedright\arraybackslash}p{0.19\textwidth}@{}}
\toprule
Gate & Decision rule & Default \\
\midrule
Search history: provenance & The search and the data used to select the claim are documented, and selection did not use the full dataset. & -- \\
Search history: multiplicity & Scalar: $p\leq\alpha/M_{\mathrm{eff}}$ in the locked direction. Brain-wide: at least one region with Benjamini--Hochberg $q\leq\alpha$ over $M_{\mathrm{eff}}$ hypotheses, in the locked direction. & $\alpha=0.05$ \\
Measured confounds & Every required covariate is declared, and no site, scanner, field-strength, or software variable is both associated with the predictor and omitted from the covariates. & $\alpha=0.05$ \\
Power & The planned design detects $\delta_{\mathrm{ref}}$, a literature value when available and $d=0.3$ otherwise, at the minimum power. & power $0.80$ \\
Analytic stability & The fraction of consistent analysis variants (forks) is at least $\rho$, over 24 forks for scalar claims and 6 for brain-wide claims. & $\rho=0.60$ \\
Replication & Scalar: locked direction and $p<\alpha$ in every replication cohort. Brain-wide: pattern correlation, replicated-region fraction, and Dice at or above the analysis-plan thresholds, with at least three shared regions and 50\% feature coverage. & $\alpha=0.05$; $0.5$, $0.5$, $0.0$ \\
\bottomrule
\end{tabular}
\end{table*}

\paragraph{Search history and multiplicity.}
The effective family size is the number of tests used for multiplicity correction. It is the largest count among the declared tests, the tests documented during claim selection, and the outcomes or regional features tested during execution. Scalar claims must reach $p\leq\alpha/M_{\mathrm{eff}}$ in the declared
direction, and brain-wide claims use Benjamini--Hochberg-adjusted $q$-values
across the regional family~\citep{benjamini1995}. A search whose selection
history is unknown, or whose selection used the full data, receives
\texttt{unverifiable\_search\_provenance}.

\paragraph{Measured confounds.}
The gate audits four measured confounds: site, scanner, field
strength, and processing-software version. For each available variable that
varies within the data, it tests association with the predictor at
$\alpha=0.05$, using a chi-square test for a group predictor, a one-way ANOVA
for a continuous predictor against a categorical variable, and a Pearson
correlation between continuous variables. It stores the test statistic,
$p$-value, and number of levels. Omitting an associated variable from the
declared covariates yields \texttt{confound\_incomplete}.

\paragraph{Statistical power.}
The power gate asks whether the planned design can detect a reference effect
$\delta_{\mathrm{ref}}$ at the minimum power set in the analysis plan. Designs below the threshold receive \texttt{under\_powered}.
Table~\ref{tab:robustness_compact} varies
$\delta_{\mathrm{ref}}$ from 0.1 to 0.5. Although the share of scalar claims
passing the power gate ranges from 17\% to 99\%, CONFIRM-Full remains at 41\%
recovery, 11\% literature false support, and no false support on controls for values from 0.2 to
0.4. Recovery changes only at $d=0.1$, falling to 14\% with no false
support.

\paragraph{Analytic stability.}
Let $\mathcal{F}$ denote the pre-specified analysis variants, or \emph{forks}, that differ in covariates, outlier handling, and model form. The default grid combines inclusion or omission of total intracranial volume and of site, three outlier rules (none, removal at
$|z|>3$, and winsorization, which caps extreme values), and ordinary least squares or a robust model, giving $|\mathcal{F}|=24$ forks. The grid is fixed before execution and
shared across claims unless an analysis plan declares another. A scalar fork is
consistent when it retains the required direction and multiplicity threshold,
and CONFIRM requires
\begin{equation}
S(C)=\frac{1}{|\mathcal{F}|}\sum_{f\in\mathcal{F}} I_f\geq\rho,
\label{eq:appendix_stability}
\end{equation}
where $I_f=1$ only for a directionally consistent, significant fork and $\rho$
is fixed in the analysis plan. Forks that fail to run or cannot estimate the specified effect remain in the denominator. Brain-wide claims must retain their regional pattern across six
forks, which combine full or minimal covariates with the three outlier rules under
ordinary least squares. A brain-wide fork is consistent when at least half of
the regions significant in the primary fit remain significant in the locked
direction.

\paragraph{Independent replication.}
The model specified by the fixed analysis plan is fitted again in every designated replication cohort.
Scalar claims must replicate with the required direction and independent
significance. Brain-wide analysis plans additionally fix thresholds for regional
overlap, replicated-region fraction, and pattern correlation. ComBat
harmonization is applied only when the analysis plan requires
it~\citep{johnson2007,fortin2018}.

\begin{table*}[t]
\centering
\small
\setlength{\tabcolsep}{6pt}
\caption{\textbf{Report-level verdicts and the reasons attached to withheld
claims.} The analysis record retains the result of every gate and every reason for withholding a claim,
even when a reporting standard displays a single verdict.}
\label{tab:appendix_verdict_taxonomy}
\begin{tabular}{@{}>{\raggedright\arraybackslash}p{0.18\textwidth}
                >{\raggedright\arraybackslash}p{0.34\textwidth}
                >{\raggedright\arraybackslash}p{0.39\textwidth}@{}}
\toprule
Verdict & Assignment rule & Example attached reasons \\
\midrule
\texttt{confirmed} & Every gate passes for the pre-specified analysis plan. & All gate states pass. \\
\texttt{fragile} & Search history or multiplicity requirements, measured-confound checks, or analytic stability fail. & \texttt{unverifiable\_search\_provenance}; \texttt{confound\_incomplete}; failed stability. \\
\texttt{under\_powered} & The planned design cannot detect $\delta_{\mathrm{ref}}$ at the required power. & Reference effect, planned sample size, and achieved power. \\
\texttt{non\_replicated} & The locked analysis plan fails in a designated replication cohort. & Missing compatible outcome, wrong sign, or non-significant effect. \\
\bottomrule
\end{tabular}

\end{table*}

\subsection{Failure Diagnosis and Candidate Validation}

A deterministic routine maps the parent analysis plan, its analysis record, and
its failed gates to a structured diagnosis,
\[
d_0=\operatorname{Diagnose}(C_0,E(C_0),\mathbf{g}(C_0,E(C_0))).
\]
The diagnosis records the primary failure, its supporting evidence, the fields
that must be preserved, and the types of revision allowed. It distinguishes
errors in the analysis plan, evidence failures, design limits, search-history
failures, and confound failures.

In round $r$, the LLM receives $C_0$, $d_0$, the list of variables available in
the search data, and a structured summary of the failures from round $r-1$. Each
candidate $C_{r,k}$ records its parent, round, question, executable analysis
plan, and a rationale connecting it to the parent. Permitted changes are
refinements of the outcome within the same modality, subgroups available in the
search data, justified measured confounds, and brain-wide patterns related to the parent claim.

The validator compares each candidate analysis plan with its parent. It rejects any change to the predictor or contrast,
direction, datasets, required gates, or existing covariates, as well as
unrelated modalities, invalid inclusion rules, unsupported numeric claims,
non-executable fields, and rewrites that change only wording. An added confound
must be available in every search-data cohort and have a scientific justification.
The agent receives invalid proposals and the rules they violated, with a limit on the number of repair attempts. Evaluation proceeds with candidates that satisfy the
schema and validation rules.

\subsection{Stopping Rules, Candidate Status, and Holdout Evaluation}

Candidates that provisionally pass on the search data are re-checked using the final count of tests. A candidate that no longer passes may return to the next round for revision. The search stops after $R$ rounds, when no valid candidate is
generated, when generation fails, or when no failed candidate remains.

Some candidates repair the parent's failure directly, for example by
adding an omitted confound. Others introduce new outcomes, subgroups, or
brain-wide patterns. Both kinds must pass every gate and receive exploratory
support on the search data.

After the search-data evaluation, each surviving candidate and its final
effective family size are frozen. A mapping fixed independently of outcomes assigns compatible analysis plans
to designated internal holdout and external cohorts.

\FloatBarrier

\section{Benchmark Construction Details}
\label{sec:appendix_benchmark_details}

\subsection{Cohorts, Modalities, and Eligibility}

NeuroClaimBench draws on 544 source items spanning structural MRI,
resting-state functional connectivity, and PET. Structural MRI for Alzheimer's
disease and aging uses ADNI~\citep{weiner2010adni} and
OASIS3~\citep{lamontagne2019oasis3}, with NACC for external
transfer~\citep{beekly2007nacc}. Normative and neurodevelopmental
functional-connectivity analysis plans use UK
Biobank~\citep{miller2016ukb,alfaroalmagro2018ukb}, HCP-family
cohorts~\citep{vanessen2013hcp,bookheimer2019hcpa}, ABCD~\citep{casey2018abcd},
ABIDE~\citep{dimartino2014abide,dimartino2017abide2}, and
ADHD-200~\citep{bellec2017adhd200}, and functional MRI for Alzheimer's disease
and aging also uses ADNI and OASIS3. Psychosis analysis plans use
COBRE~\citep{aine2017cobre}, FBIRN~\citep{keator2016fbirn}, ChineseSZ,
BSNIP2~\citep{bsnip2nda,clementz2022bsnip}, Olin, and Johns Hopkins data, with
CNP for external transfer~\citep{poldrack2016cnp}.

Each claim is tied to its cohort pair. An analysis plan is executable only when
every named cohort contains its outcome, predictor, covariates, and group
definition, and the model can estimate the specified effect. Low-power designs remain eligible for evaluation by the power gate. Differences in scanner, site, and
processing remain part of the replication challenge.

Each benchmark case records three fields: the split
(scientific, external-transfer, or designed control), the reference label (support, withhold, or unresolved), and the observed outcome
(confirmed, withheld, error, or not evaluated). Literature evidence, model agreement, the source of the question, and
reference strength are recorded for sensitivity analyses.

Each claim records its question, analysis plan, dataset roles, and evidence
sources, with hashes identifying the frozen question, plan, and data partitions.
Controls also record their assignments, selected rows, generation parameters,
and random seeds. Items with matching executable plans and scientific content
are merged, but distinct random assignments remain separate tests.

\subsection{Source Corpus}

The question corpus contains 291 questions: 41 literature-derived questions
with executable cohort pairs and 250 questions independently proposed by
GPT-5.5. The generated questions comprise 50 in each of five population groups:
normative fMRI, ADHD, ASD, AD/aging, and psychosis. These questions yielded
289 executable analysis plans, all evaluated without execution errors.

NeuroClaimBench combines these sources with additional scientific claims,
external-transfer tasks, and designed controls. After merging duplicate items,
the source corpus contains 544 items: 332 scientific, 62 external-transfer, and
150 designed controls. Excluding 40 ambiguous mappings and 15 non-executable
items leaves 489 executable claims: 279 scientific, 60 external-transfer, and
150 designed controls.

\begin{table*}[t]
\centering
\small
\setlength{\tabcolsep}{6pt}
\caption{\textbf{NeuroClaimBench composition.} The Support, Withhold, and Unresolved columns give reference labels among
executable claims. Scored metrics use the labeled scientific claims and the designed controls.}
\label{tab:ncb_v21_composition}
\begin{tabular}{@{}lrrrrr@{}}
\toprule
Split & Source & Executable & Support & Withhold & Unresolved \\
\midrule
Scientific & 332 & 279 & 51 & 19 & 209 \\
External-transfer & 62 & 60 & 18 & 30 & 12 \\
Designed controls & 150 & 150 & 0 & 150 & 0 \\
\midrule
Total & 544 & 489 & 69 & 199 & 221 \\
\bottomrule
\end{tabular}

\end{table*}

\subsection{Outcome-Blind Eligibility Assessment}

Corrections that aligned an analysis plan with its question kept the
question's predictor, contrast, population, outcome, modality, and direction,
and could also add measured confounds as covariates. Before execution, deterministic checks verify the groups, outcomes, required columns, design rank, and residual
degrees of freedom. Gemini 3.5 Flash reviewed eligibility without access to results for 362 literature-derived records and marked 40 otherwise aligned
records as ambiguous, which remain unresolved and excluded from all reported metrics. Designed controls were validated deterministically. The prompts
excluded CONFIRM outcomes, $p$-values, effect estimates, existing labels, and
follow-up results.

\subsection{Reference Construction and Coverage}
\label{sec:appendix_reference_construction}

GPT-5.5 and Gemini 3.5 Flash assessed fixed sets of PubMed evidence, and Claude
Opus 4.8 served as the independent arbitrator. Every cited PMID or DOI had to
appear in the fixed evidence set, and the cited study had to match the claim's population, contrast, imaging modality, outcome, and direction. Of the 96
score-eligible literature references, 44 had agreement among all three models,
15 had arbitrator-plus-assessor agreement, and 37 had assessor-only agreement.
The 47 strict references comprise 39 all-three and 8
arbitrator-plus-assessor cases, and the 49 provisional references comprise 5,
7, and 37 cases of these three types. Neuroimaging experts reviewed and finalized all literature labels.

The 150 controls in the control set and 22 random-label controls in the external-transfer split receive negative labels from their design. They are separate from the strict literature references. Evidence was insufficient for 276 source items, including 221
executable unresolved items. For these items, no article met the requirement that
two independent models judge it to provide direct evidence for the exact frozen
contrast.

\begin{figure*}[t]
\centering
\includegraphics[width=\textwidth]{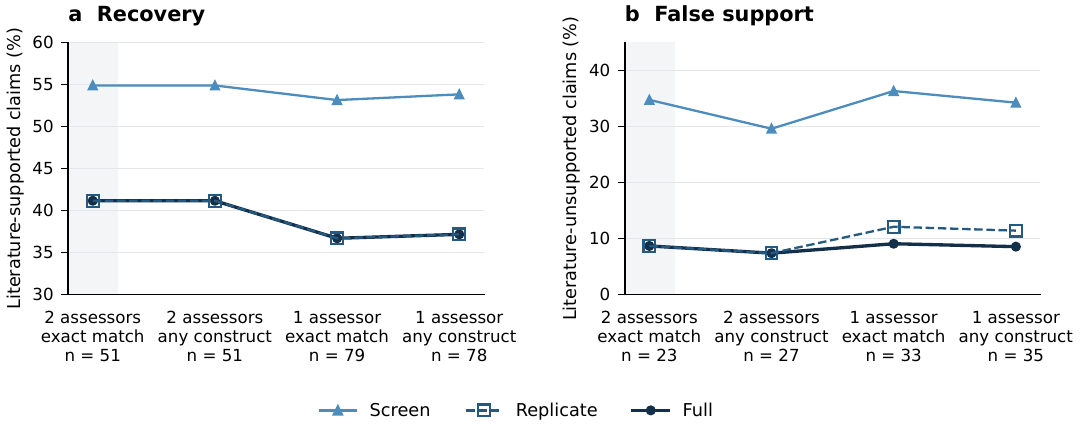}
\caption{\textbf{Sensitivity to the requirements for literature reference labels.} A reference
requires two assessors to identify direct evidence for the exact frozen
contrast, and 209 executable scientific claims fail this criterion. Reference rules progress from strictest to most permissive, and $n$ is the number of labeled claims under each rule. Shading marks the two-assessor, exact-match rule. Open squares and filled circles distinguish coincident Replicate and Full results. CONFIRM's verdicts are fixed, so only the labeled
set changes. \textbf{a}, Recovery. \textbf{b}, False support, including four additional literature-unsupported references. The ordering Full $\leq$
Replicate $<$ Screen holds under every reference rule. }
\label{fig:reference_bar_sensitivity}
\end{figure*}

We reapplied the recorded assessor votes under three less demanding requirements for assigning reference labels
(Figure~\ref{fig:reference_bar_sensitivity}). Allowing two assessors to use studies that only partly match the claim added four literature-unsupported references without
changing recovery. Accepting a single assessor added 42 labels without
cross-model agreement, below the agreement required for provisional references. The ordering of false support was unchanged under every rule. CONFIRM-Full false support ranged from 7.4\% to 9.1\%, and
recovery fell by 1.7 points for CONFIRM-Screen and by 4.5 points for the stricter
standards, with CONFIRM-Full moving from 41\% to 37\%.

\FloatBarrier

\section{Experimental Details and Additional Results}
\label{sec:appendix_experiment_details}

\begin{table*}[t]
\centering
\small
\setlength{\tabcolsep}{5pt}
\caption{\textbf{Claim-evaluation comparison and CONFIRM-Full scope.} In the
matched comparison, every method scores the same 31 literature-supported and 15
literature-unsupported claims from the common scalar set and the same 150
designed controls. The last row reports CONFIRM-Full on the strict references of the full literature set, 23 supported and 3 unsupported. Values are percentages of claims supported. Bracketed 95\% confidence intervals accompany the matched-comparison rows.
Arrows show whether higher ($\uparrow$) or lower ($\downarrow$) is better.}
\label{tab:claim_evaluation_common_scalar}
\begin{tabular}{@{}lrrr@{}}
\toprule
Evaluation & Recovery $\uparrow$ & Literature false support $\downarrow$ & Control false support $\downarrow$ \\
\midrule
\multicolumn{4}{@{}l}{\textit{Matched common-scalar comparison}}\\
VERITAS-adapted & 65 [45, 81] & 40 [16, 68] & 38 [30, 46] \\
NeuroClaw-adapted & 48 [30, 67] & 13 [2, 40] & 23 [16, 30] \\
Direct LLM judge & 55 [36, 73] & 20 [4, 48] & 29 [22, 37] \\
Significance filter & 52 [33, 70] & 13 [2, 40] & 28 [21, 36] \\
CONFIRM-Screen & 61 [42, 78] & 40 [16, 68] & 3 [1, 8] \\
CONFIRM-Replicate & 48 [30, 67] & 13 [2, 40] & 1 [0, 4] \\
CONFIRM-Full & 48 [30, 67] & 13 [2, 40] & 0 [0, 2] \\
\addlinespace
\multicolumn{4}{@{}l}{\textit{CONFIRM-Full on the full literature set}}\\
Scientific, all modalities, strict only & 48 & 0 & -- \\
\bottomrule
\end{tabular}

\end{table*}

\begin{table*}[t]
\centering
\small
\setlength{\tabcolsep}{3.5pt}
\caption{\textbf{Effect of removing each gate from CONFIRM-Full.} Each row reports the increase in support, in percentage points ($\Delta$), when one gate is omitted and all other thresholds are held fixed. Recovery is computed over 51 literature-supported references, literature false support over 19 literature-unsupported references, and control false support over 150 designed controls.}
\label{tab:confirm_robustness}
\begin{tabular}{@{}lrrr@{}}
\toprule
Removed gate & $\Delta$ literature recovery & $\Delta$ literature false support & $\Delta$ control false support \\
\midrule
Search history: provenance & +0 & +0 & +0 \\
Search history: multiplicity & +0 & +0 & +0 \\
Measured confounds & +2 & +0 & +24 \\
Power & +0 & +0 & +1 \\
Analytic stability & +0 & +0 & +0 \\
Replication & +8 & +26 & +2 \\
\bottomrule
\end{tabular}

\end{table*}

\begin{figure}[t]
\centering
\newcommand{\policycomparisoncaption}{\textbf{Recovery and false support under each reporting standard.}
Results on the literature set, for the direct LLM judge and the three CONFIRM standards. Green bars show recovery and red bars show literature false support.}
\ifvenueaamas
\includegraphics[width=\smallfigwidth]{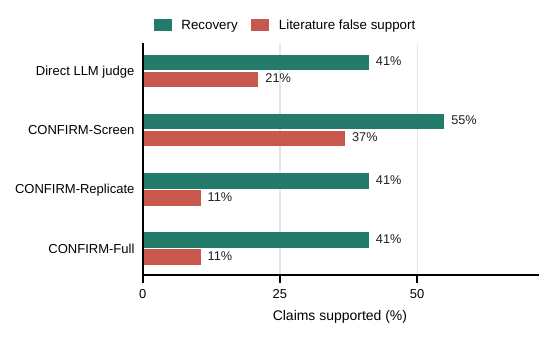}
\caption{\policycomparisoncaption}
\label{fig:confirm_policy_comparison}
\else
\begin{minipage}[c]{0.54\textwidth}\centering
\includegraphics[width=\linewidth]{figures/fig_confirm_policy_comparison.pdf}
\end{minipage}\hfill
\begin{minipage}[c]{0.42\textwidth}
\caption{\policycomparisoncaption}
\label{fig:confirm_policy_comparison}
\end{minipage}
\fi
\end{figure}

\subsection{Comparison Protocol}
\label{sec:appendix_comparison_protocol}

The three reporting standards use the same analysis plans, gate thresholds,
and gate results. They differ in the gates required for support, with
CONFIRM-Full requiring all five. Because the
significance filter has no decision rule for brain-wide claims, comparisons that
include it use the common scalar set. Removing one gate at a time measures its contribution to the verdict.

The direct LLM judge receives the frozen claim and the evidence with reference labels excluded in a single GPT-5.5 call, whereas NeuroClaw-adapted makes three calls per
claim. The judge applies no aggregation rule for brain-wide claims. The
significance filter requires an unadjusted $p<0.05$ in the locked direction in
discovery and in every replication cohort, with no multiplicity, confound,
power, or stability requirement. In follow-up search, failure-blind retry
receives only the prior failure, whereas Self-Refine~\citep{madaan2023selfrefine}
adds a critique call before each refinement, which increases the number of LLM calls for the same search budget.

The 150 designed controls comprise 42 site-confounded, 42 broad-search, 42
random-label, 21 underpowered, and 3 nonreplicating cases. Site-confounded
controls account for 74\% of the controls that VERITAS-adapted supports, 95\%
for the direct judge, 98\% for the significance filter, and 97\% for
NeuroClaw-adapted. VERITAS-adapted also accepts 24\% of broad-search controls,
because it relies on discovery evidence alone. CONFIRM-Full rejects every
control in all five families.

\FloatBarrier

\subsection{Why CONFIRM Withholds Literature Claims}
\label{sec:appendix_withheld}

\begin{table}[htbp]
\centering
\footnotesize
\setlength{\tabcolsep}{3pt}
\caption{\textbf{Gates failed by the literature claims that CONFIRM-Full withholds (\%).} Withheld is the share of each set that CONFIRM-Full withholds, out of 51 literature-supported and 19 literature-unsupported claims. The remaining columns give the share of withheld claims that fail replication (Repl.), multiplicity (Mult.), stability (Stab.), power, and measured confounds (Conf.). A claim can fail several gates, so the rows do not sum to 100.}
\label{tab:withheld_gates}
\begin{tabular}{@{}lrrrrrr@{}}
\toprule
Label & Withheld & Repl. & Mult. & Stab. & Power & Conf. \\
\midrule
Supported & 59 & 87 & 73 & 73 & 20 & 3 \\
Unsupported & 89 & 94 & 71 & 71 & 0 & 0 \\
\bottomrule
\end{tabular}
\end{table}

Table~\ref{tab:withheld_gates} shows why CONFIRM-Full withholds literature claims on the full literature set. Most withheld claims, whether or not the literature supports them, fail to replicate in the independent cohort, and many also fail the multiplicity or stability gate. Power and confound failures are rare among these claims.

\subsection{Holdout Support for Standalone Verdicts}
\label{sec:appendix_heldout_verdicts}

CONFIRM-Full confirmed 74 of 289 standalone analysis plans and withheld 215 (74\%). We reran all five gates for these frozen plans on designated holdout discovery and replication partitions excluded from the initial verdicts.

The audit included 65 confirmed and 144 withheld claims.

On holdout, 25\% of initially confirmed claims and 3\% of withheld claims passed Full, and 63\% and 15\%, respectively, passed the replication gate (Table~\ref{tab:heldout_standards}). Power failures explain much of the difference between passing Full and passing replication alone: 39 of 65 initially confirmed claims failed the power gate, including 22 that passed every other gate.

\begin{table}[htbp]
\centering
\footnotesize
\setlength{\tabcolsep}{3pt}
\caption{{\textbf{Holdout outcomes by initial verdict.} Entries count claims. A claim can fail several gates.}}
\label{tab:heldout_standards}
\begin{tabular}{@{}lrr@{}}
\toprule
Holdout outcome & Confirmed & Withheld \\
\midrule
Evaluated claims & 65 & 144 \\
Full standard passed & 16 & 5 \\
Replication gate passed & 41 & 22 \\
\midrule
\multicolumn{3}{@{}l}{Failed gates (overlapping)} \\
Search history / multiplicity & 6 & 124 \\
Measured confounds & 0 & 4 \\
Power & 39 & 72 \\
Analytic stability & 6 & 126 \\
Independent replication & 24 & 122 \\
\bottomrule
\end{tabular}
\end{table}

\subsection{Diagnosis Feedback for an Existing Agent}
\label{sec:appendix_neuroclaw_feedback}

NeuroClaw-adapted writes
questions and analysis plans, CONFIRM evaluates the 280 executable drafts, and the
200 withheld claims are revised using one of two types of feedback with the same budget: self-critique, in which the agent critiques its supported or withheld verdict, or CONFIRM's gate diagnosis
(Table~\ref{tab:neuroclaw_feedback}). Diagnosis finds supported follow-ups for
more withheld claims than self-critique while using fewer calls, because
self-critique adds a separate critique call in each round. Every reported
candidate is re-checked on holdout partitions withheld from the search, where
19\% of diagnosis candidates pass, against 5\% of self-critique candidates. No method supports any designed control.

\begin{table*}[t]
\centering
\small
\setlength{\tabcolsep}{6pt}
\caption{\textbf{Diagnosis feedback for an existing agent.} NeuroClaw-adapted
drafts 280 executable claims, CONFIRM evaluates them, and the agent uses feedback to revise only the
200 withheld claims. Reported and Holdout are percentages of the 280 executed
drafts, Added is the percentage of withheld claims revised into supported
claims, and Control is false support on the 42 site-confounded controls. Both arms share the model, the budget of $R=3$ rounds and $K=5$
candidates per round, the validator, and the multiplicity correction. Holdout
support uses partitions excluded from revision. Arrows show whether higher
($\uparrow$) or lower ($\downarrow$) is better.}
\label{tab:neuroclaw_feedback}
\begin{tabular}{@{}lrrrrr@{}}
\toprule
Regime & Reported (\%) $\uparrow$ & Added (\%) $\uparrow$ & Calls $\downarrow$ & Holdout (\%) $\uparrow$ & Control (\%) $\downarrow$ \\
\midrule
No feedback & 28.6 & n/a & n/a & 5.7 & 0 \\
Self-critique & 35.4 & 9.5 & 1,357 & 6.4 & 0 \\
CONFIRM diagnosis & 37.5 & 12.5 & 894 & 8.9 & 0 \\
\bottomrule
\end{tabular}

\end{table*}

NeuroClaw-adapted generates its own questions from the same executable variables, with the same number of questions
per population group, as in standalone mode.

Control evaluation uses the 42 frozen site-confounded benchmark cases.
Supplying their existing analysis plans preserves the omission of
\texttt{site} that creates the deliberately introduced confound. This family tests the
measured-confound gate, and each baseline accepts cases from it
(Figure~\ref{fig:confirm_frontier}b).

Holdout evaluation maps each claim to its predeclared cohort pair and reruns
the unchanged gates.

\begin{table*}[t]
\centering
\small
\setlength{\tabcolsep}{6pt}
\caption{\textbf{Follow-up search runs.} The first three arms revise the 215 claims that CONFIRM-Full withheld, with $R=3$ rounds and up to $K=5$ candidates per round, and differ only in the feedback given to the LLM. The last two arms give CONFIRM's diagnosis for all 150 designed controls, with a small and a large budget. Generated counts candidate analysis plans proposed by the LLM, Retained counts distinct candidates kept after removing duplicates, and Evaluated counts candidates that passed validation and were run. Supported counts candidates that
pass every gate on the search data after correcting for all analyses tried.
Parents is the percentage of starting claims with at least one
supported candidate.}
\label{tab:feedback_control}
\begin{tabular}{@{}lrrrrr@{}}
\toprule
Arm & Generated & Retained & Evaluated & Supported & Parents (\%) \\
\midrule
CONFIRM diagnosis, $R=3$, $K=5$ & 3,187 & 3,059 & 2,971 & 70 & 11 \\
Failure-blind retry, $R=3$, $K=5$ & 3,275 & 3,178 & 3,073 & 49 & 9 \\
Self-Refine, $R=3$, $K=5$ & 3,259 & 2,361 & 2,161 & 54 & 11 \\
Designed controls, $R=1$, $K=2$ & 334 & 296 & 293 & 0 & 0 \\
Designed controls, $R=10$, $K=10$ & 13,371 & 12,313 & 10,716 & 0 & 0 \\
\bottomrule
\end{tabular}

\end{table*}

\subsection{Worked Example: Revising One Withheld Claim}
\label{sec:appendix_worked_example}

Figure~\ref{fig:worked_example} traces one withheld claim through the follow-up
search of Section~\ref{sec:feedback_results}.

\paragraph{Step 1: the original claim and its verdict.}
The claim states that, among cognitively normal OASIS3 participants, older age
is associated with larger ventricular volume. The analysis plan adjusts for sex,
estimated total intracranial volume (eTIV), and site, and it replicates the claim in a separate
OASIS3 partition reserved for replication. CONFIRM withholds the claim at the
measured-confound gate. OASIS3 scans were processed with four versions of
FreeSurfer, and the version used is associated with participant age. Because
the analysis omits the software version, software changes could account for part of the apparent age effect. CONFIRM therefore labels
the claim \texttt{fragile}.

\paragraph{Step 2: the diagnosis given to the LLM.}
CONFIRM identifies the omitted measured confound and lists the fields that
each revision must preserve: the scientific question, the outcome family, the
expected direction, the datasets, and every gate threshold.

\paragraph{Step 3: the LLM's proposals.}
Guided by this diagnosis, the LLM proposes three revisions. Proposal (a) adds
the FreeSurfer version as a covariate, directly repairing the diagnosed failure.
Proposal (b) adds the software version and restricts the sample to participants
aged 65 or older. Proposal (c) adds the software version and tests a six-region
structural pattern instead of ventricular volume alone.

\paragraph{Step 4: validation.}
Before execution, the deterministic validator compares each proposal with the
parent analysis plan. Proposals (a) and (c) preserve every required field and
pass validation. Proposal (b) is rejected without being run, because the age cutoff
is not among the subgroups offered to the LLM.

\paragraph{Step 5: evaluation on the search data.}
The two valid proposals run on the same discovery and replication data as the
parent and must pass every gate. Each result is corrected for every analysis
tried during the search, and both (a) and
(c) pass every gate after this correction. Their support remains exploratory at
this stage, because the search used the same data.

\paragraph{Step 6: evaluation on holdout data.}
Finally, (a) and (c) are tested on holdout data never used during the search,
and both are supported there. The parent claim, which omits the software
version, is not supported on the same holdout data.

\subsection{Follow-Up Search Runs and Designed-Control Stress Tests}
\label{sec:appendix_followup_runs}

Table~\ref{tab:feedback_control} reports candidate and parent counts for the
matched runs with $R=3$ rounds and $K=5$ candidates per round and for two designed-control runs. All runs use the search data only. The two designed-control arms start from all 150 designed controls and give the LLM CONFIRM's diagnosis, with $R=1$, $K=2$ and $R=10$, $K=10$. In the high-budget arm, none of the 10,716 evaluated candidates passes
even provisionally, and no execution or identifiability failure occurs.

\FloatBarrier

\section{Scope and Limitations}
\label{sec:limitations}

NeuroClaimBench covers a finite set of executable claims that represents some target populations more than others. Its
reference labels come from expert review of the retrieved evidence, its
designed controls target specific gates, and it covers three imaging modalities.

The gates assess the recorded analyses and measured confounds under fixed thresholds. Passing them does not establish biological truth.

\end{document}